# Enabling ASR for Low-Resource Languages: A Comprehensive Dataset Creation Approach

*CaseStudy: Armenian Language Audiobooks

Ara Yeroyan
*Data Science Department*
*American University of Armenia*
Yerevan, Armenia
ara_yeroyan@edu.aua.am

Nikolay Karpov
*NeMo Conversational AI*
*Nvidia*
Yerevan, Armenia
nkarpov@nvidia.com

***Abstract*—In recent years, automatic speech recognition (ASR) systems have significantly improved, especially in languages with a vast amount of transcribed speech data. However, ASR systems tend to perform poorly for low-resource languages with fewer resources, such as minority and regional languages. This study introduces a novel pipeline designed to generate ASR training datasets from audiobooks, which typically feature a single transcript associated with hours-long audios. The common structure of these audiobooks poses a unique challenge due to the extensive length of audio segments, whereas optimal ASR training requires segments ranging from 3 to 15 seconds. To address this, we propose a method for effectively aligning audio with its corresponding text and segmenting it into lengths suitable for ASR training. Our approach simplifies data preparation for ASR systems in low-resource languages and demonstrates its application through a case study involving the Armenian language. Our method, which is "portable" to many low-resource languages, not only mitigates the issue of data scarcity but also enhances the performance of ASR models for underrepresented languages.**

## I. Introduction

Automatic Speech Recognition (ASR) is a technological innovation designed to transform spoken language into written text. The core functionality of ASR systems involves capturing verbal input and converting it into accurate textual representations, primarily measured by the Word Error Rate (WER) [1]. This capability is crucial for facilitating effective communication across both human-to-human and human-to-machine interfaces. It has been integrated into a multitude of applications, enhancing accessibility and automation across various domains. These applications range from air traffic control and biometric security to more consumer-oriented uses such as closed captioning for digital media, voice message transcription, and smart home automation systems. The implementation of ASR technology has evolved significantly, becoming more sophisticated as its applications widen [2].

ASR technology has achieved remarkable advancements in high-resource languages such as English due to the abundance of available data. These systems benefit immensely from extensive databases of paired audio-text samples, allowing for more refined training and accuracy. In contrast, the performance of ASR systems in low-resource languages—often minority, regional, or dialectical—lags significantly. The primary challenge lies in the scarcity of annotated speech datasets. Without substantial data, training effective ASR systems for these languages becomes exceedingly difficult. Traditional strategies to mitigate these data limitations include leveraging high-resource languages for transfer learning, employing unsupervised pre-training techniques with unpaired data, and utilizing synthetic data augmentation to enhance model robustness and prevent overfitting [3]. However, despite its potential, transferring knowledge to low-resource languages is computationally demanding and prone to catastrophic forgetting [4].

Given these challenges and the limited effectiveness of current methods in improving ASR for low-resource languages, this study proposes a novel pipeline designed to create audio book datasets. This method leverages the typical structure of audio books—a single transcript associated with long-duration audios (per chapter for example)—to produce dataset suitable for training robust ASR systems. By focusing on the available open-source data in the low-resource Armenian language, we propose a data processing pipeline, obtain a baseline model, and use it to align audio books and retrain on an extended dataset. We also discuss the training and alignment limitations and propose solutions that can be extended for many scenarios, such as low training or data resources available, thereby supporting broader efforts to enhance ASR capabilities for underrepresented languages globally.

## II. Data Overview

### A. *General ASR supervised data schema*

ASR datasets typically consists of a collection of audio files paired with their corresponding ***normalized***[1] transcriptions. The audio samples are short (between 3 to 15 seconds) enough to capture acoustic information and recognize distinct speech sounds without loosing the context while meeting computational demands for training. These audio files are typically segmented into complete phrases or sentences, ensuring coherent speech units for effective ASR training. Unfortunately, training high-quality ASR models require extensive data, often hundreds of hours, to cover various speech nuances and to perform reliably across different speaking styles and environments.

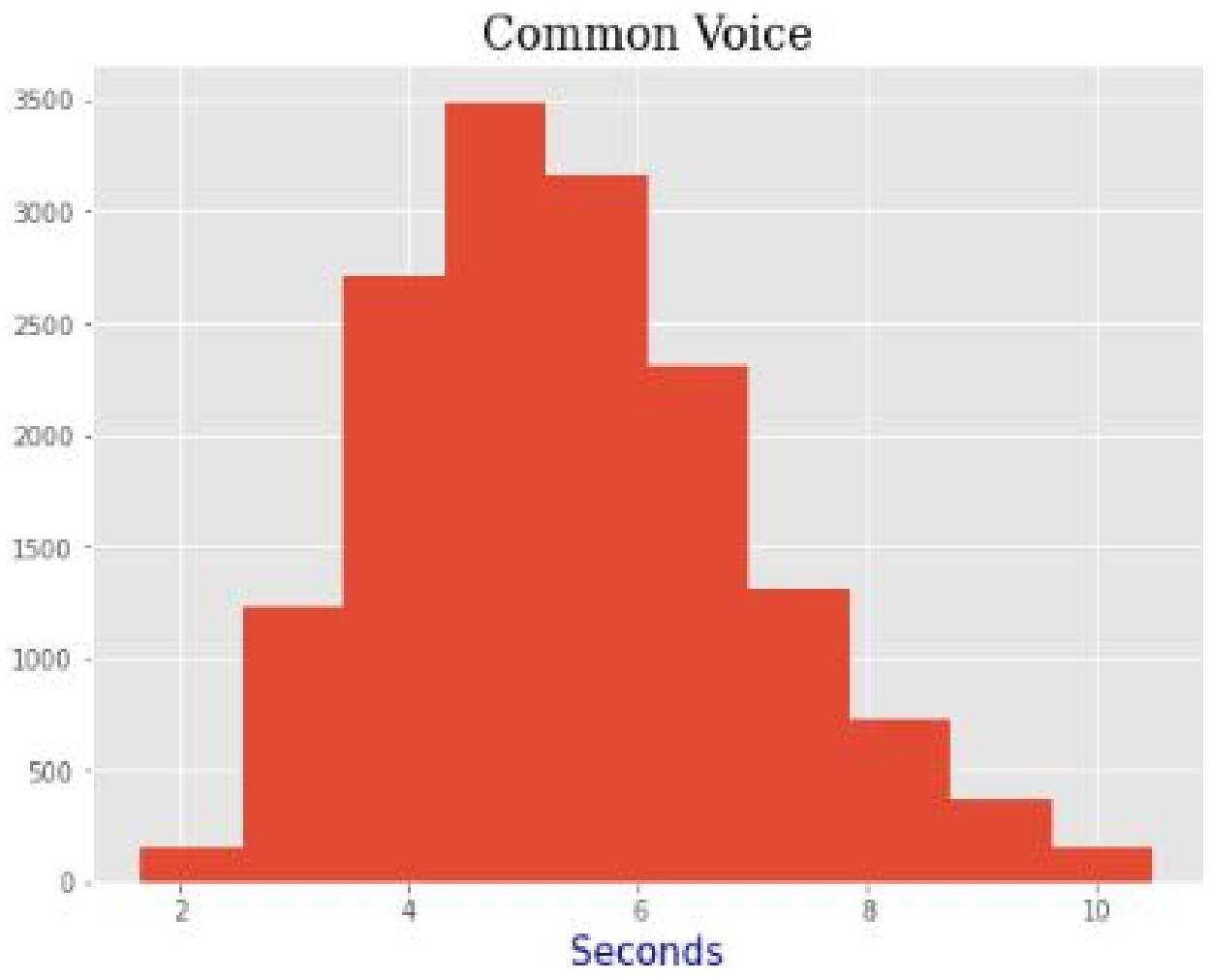

Fig. 1: MCV durations

[1]Standardise text units such as: numbers, abbreviations, URLs, etc. - NeMo-text-processing

### B. *Mozilla Common Voice (MCV)*

MCV [6] offers a multi-language dataset for ASR development, including underrepresented languages like Armenian. It is curated through crowd-sourcing, allowing for a diverse representation of accents and dialects (see Table I). As of version 17.0*, it comprises over 23 hours of **validated** audio samples from numerous speakers. MCV datasets are also good as audios typically consists of one full ***normalized**** sentence, vital for ASR training (see Fig 1). It also comes with default TRAIN, DEV and TEST sets along with annotation of validated audio recordings. Thus, this dataset, with its open licensing, serves as an essential resource for training and bench-marking ASR models.

### C. *Audiobooks*

Grqaser [7], akin to projects such as Gutenberg for English, provides a set of Armenian audio-books, analogous to collections available in various other low-resource languages. However, the data configuration presented by these resources ***diverges from the desired format*** (see Fig 2) for ASR training as outlined in Section II-A.

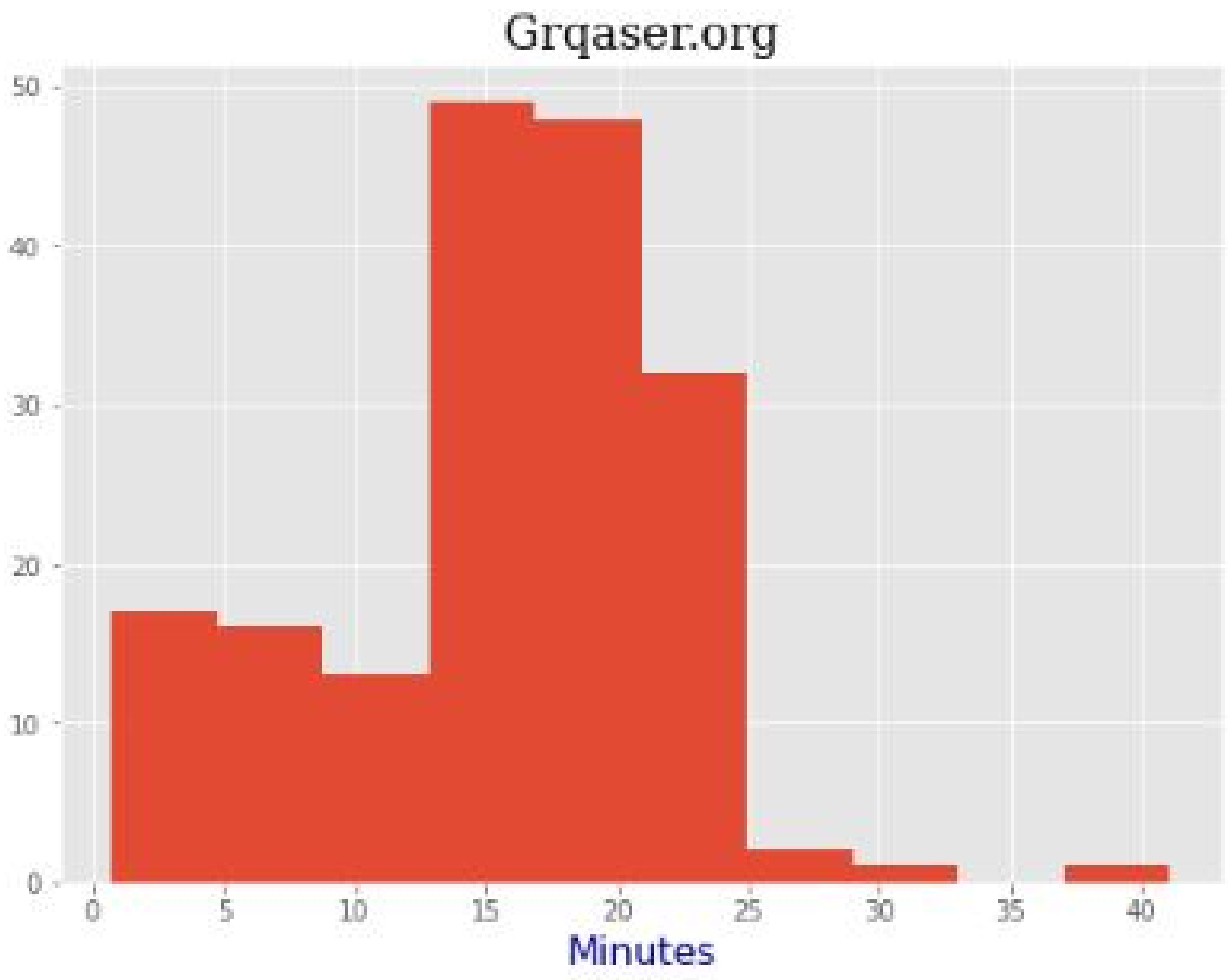

Fig. 2: Grqaser durations (8 audio books)

Take, for example, the book ”FOR THE SAKE OF HONOR”, which has single transcript alongside

| Path | Sentence | Up Votes | Age | Gender | Accent | Duration(s) |
|---|---|---|---|---|---|---|
| common_voice_hy-AM_39459841.mp3 | Մետաղական քրոմն... | 2 | twenties | female | eastern | 7.56 |
| common_voice_hy-AM_39459843.mp3 | Դա այն տեղանքի... | 4 | thirties | male | eastern | 5.940 |
| common_voice_hy-AM_39459845.mp3 | Մարոկկոյի քաան... | 4 | thirties | female | no accent | 9.216 |
| common_voice_hy-AM_39459850.mp3 | Նրանք փրկվել են... | 4 | twenties | female | eastern | 3.708 |
| common_voice_hy-AM_39459896.mp3 | Իրականում չկա... | 4 | thirties | male | Native | 5.220 |

TABLE I: Sample Data Entries from the Mozilla Common Voice Armenian Dataset.

37 audio segments, each ranging from a few minutes to 20minutes, cumulatively spanning 3.7 hours (see Fig 3).

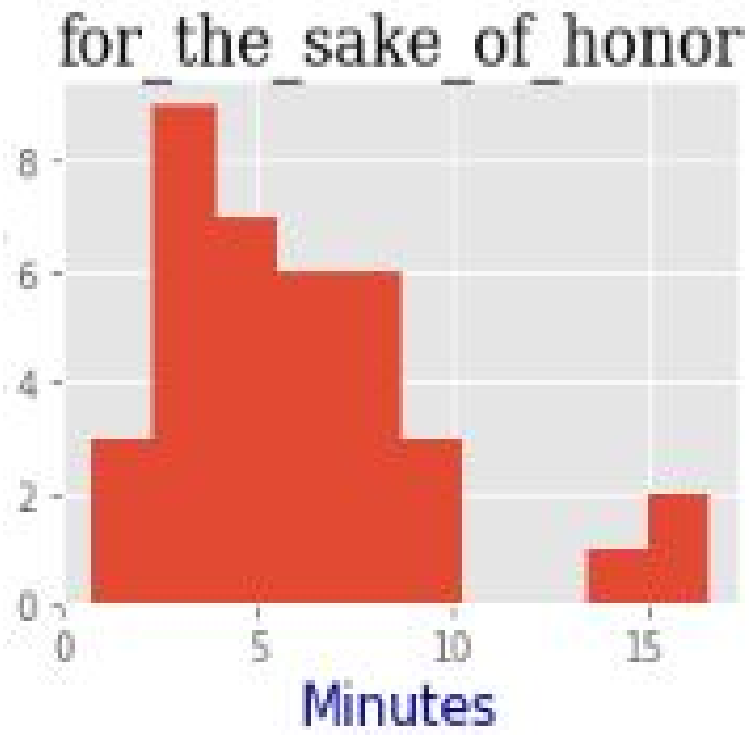

Fig. 3: Durations of ”FOR THE SAKE OF HONOR” audios - https://grqaser.org/am/713/?c=22

Utilization of such dataset necessitates a text normalization, then a 2 phase process: alignment and segmentation. Firstly, aligning the multiple audios to the monolithic transcript—potentially via ***manual inspection****. Then, cutting the audios into shorter, model-digestible segments—potentially via **NFA**[2]. This segmentation, a non-trivial task plagued by the need for having complete utterances in the short audio cuts, poses a significant challenge in data preparation for ASR model training.

***Manual Inspection***: We have a single transcript containing the text of sequential records of audios (e.g. single audio per chapter of the book). So per book processing is - begin with the first audio, listening until the end and using a text search function (CTRL+F) to find the corresponding ending words in the transcript. Align this text segment with the audio. Repeat this process consistently for each subsequent file, making sure the start of each new segment aligns with the end of the previous. A medium level intensive task repeated for all the audios (37 our example: ”FOR THE SAKE OF HONOR”).

## III. Data Processing

During our study, the primary source in terms of quality and size was MCV, specifically Common Voice Corpus 16.1 (January 5, 2024). In contrast to current v17 release, it was almost half of its size - consisting a total of 24 hours of audio. Thus, the majority of our data processing and trainings we conducted on this smaller dataset to obtain a POC - intended to be used on the v17 release (see Model Training for more details on this approach). So, this v16 corpus is divided into the following data splits (not mutually exclusive):

| Validated Data | Hours |
|---|---|
| Train | 5.16 |
| Dev | 3.83 |
| Test | 4.31 |

| Data Category | Hours |
|---|---|
| Other | 8.94 |
| Validated | 13.39 |
| Invalidated | 1.28 |

TABLE II: Distribution of Data in Mozilla Common Voice Corpus 16.1

Given the scarcity of data, we made the decision to disregard the **”validated”** status and combined the TRAIN and DEV sets with the OTHER category, filtering out any repetitions. This strategy resulted in a consolidated training corpus of approximately 18 hours. Since we merged the DEV set into the training data, we used the 6-hour-long test set for validation in our experiments. While unconventional, this approach can be justified for conducting numerous short experiments aimed at hyper-parameter optimization, as discussed in the next chapter.

The next step is to preprocess the corpus, both the texts and audios. The common steps performed by preprocessing pipeline[3] can be categorized into 2 sections:

### A. Data Cleaning

- **Normalize Texts:** While the MCV data is already normalized, for others this might not be the case
- **Resampling Audios:** A sampling rate of 16 kHz is generally sufficient for Automatic Speech Recognition (ASR) purposes
- **Language-Specific Grammatical Rules:** Enforce **regex** to map punctuation, combine variations of dashes, handle direct/indirect speech punctuation, and perform character transformations (e.g. **”ó”** to **”o”**).
- **Removing Extra Symbols:** Certain symbols such as `"<<"`, `">>"`, `"("`, and `")"` may not contribute meaningfully to ASR and can be removed.
- **Dropping Non-Alphabetic Sentences:** Containing characters outside the alphabet (e.g. English letters, or UTF artifacts) are dropped to maintain linguistic relevance.
- **Dropping Corrupted Audios:** Audios that are corrupted or of poor quality (e.g. extremely low Zero-Passing rate or audios containing many silences) are removed from the dataset

[2]Aligns audio timestamps with corresponding texts -nfa$_{a}ligner$

[3]NeMo-SDP - provides various speech/text processors

### B. Data Preparation

- **Training a Tokenizer:** A tokenizer specific to the given language is trained. For instance, a SentencePiece Unigram tokenizer (spe unigram) is used for our study.
- **Creating a NeMo-Format Manifest File:** This file typically includes fields such as **audio_filepath**, **text**, **duration** ..., providing essential metadata for training.
- **Tarring Audios into Sharded Buckets:** Audios are packaged into sharded buckets based on their durations. This method, prefered by **webdataset**, enhances batching strategy and facilitates faster training.

## IV. Model Training

After applying these preprocessing steps to the MCV v16 dataset, we proceeded to fine-tuning **stt_en_conformer_ctc_large** conformer checkpoint, to obtain a baseline model (*see Ablation IX-B*). For this purpose, we utilized the Encoder-Decoder based Connectionist Temporal Classification (**CTC**) Byte-Pair-Encoding (**BPE**) model. This model was chosen due to its effectiveness in handling sequence-to-sequence tasks like ASR. Specifically, the CTC loss employed indirectly results model to handle variable-length output sequences and is robust to alignment errors. Moreover, as Conformer-CTC architecture (Fig 4) combines the strengths of self-attention and convolutional modules, it can effectively capture both global interactions and local correlations within the input data. In addition, the use of CTC loss and decoding makes it a non-auto-regressive model, simplifying the training process and enabling efficient inference.

Training on the 18-hour-long corpus typically required around 2 days to achieve a **WER** close to **0.3**. In our training approach, we adopted a strategy of conducting multiple short training experiments (up to 70 epochs) to explore various hyper-parameter configurations. The goal was to identify an optimal set of hyper-parameters that could then be applied to larger datasets (we were expecting the **v17** release after organising its collection) for a prolonged duration of training to establish a solid baseline.

Ultimately, the insights gained from these short training experiments (see ablation study for more details) helped us to train a baseline model on the extended 38hours longs MCV **v17** dataset reaching **0.19 WER** without and **0.15 WER** with post-processing[4] on 6.75 hours long test set. When ignoring punctuation, WER becomes **0.124**! As anything below **0.2 WER** is commonly accepted as good ASR model, we could now use the obtained baseline for aligning the previously mentioned Audio Books dataset. Considering the absence of open-source Armenian ASR models demonstrating comparable performance, we made our model available on the Hugging Face platform[5].

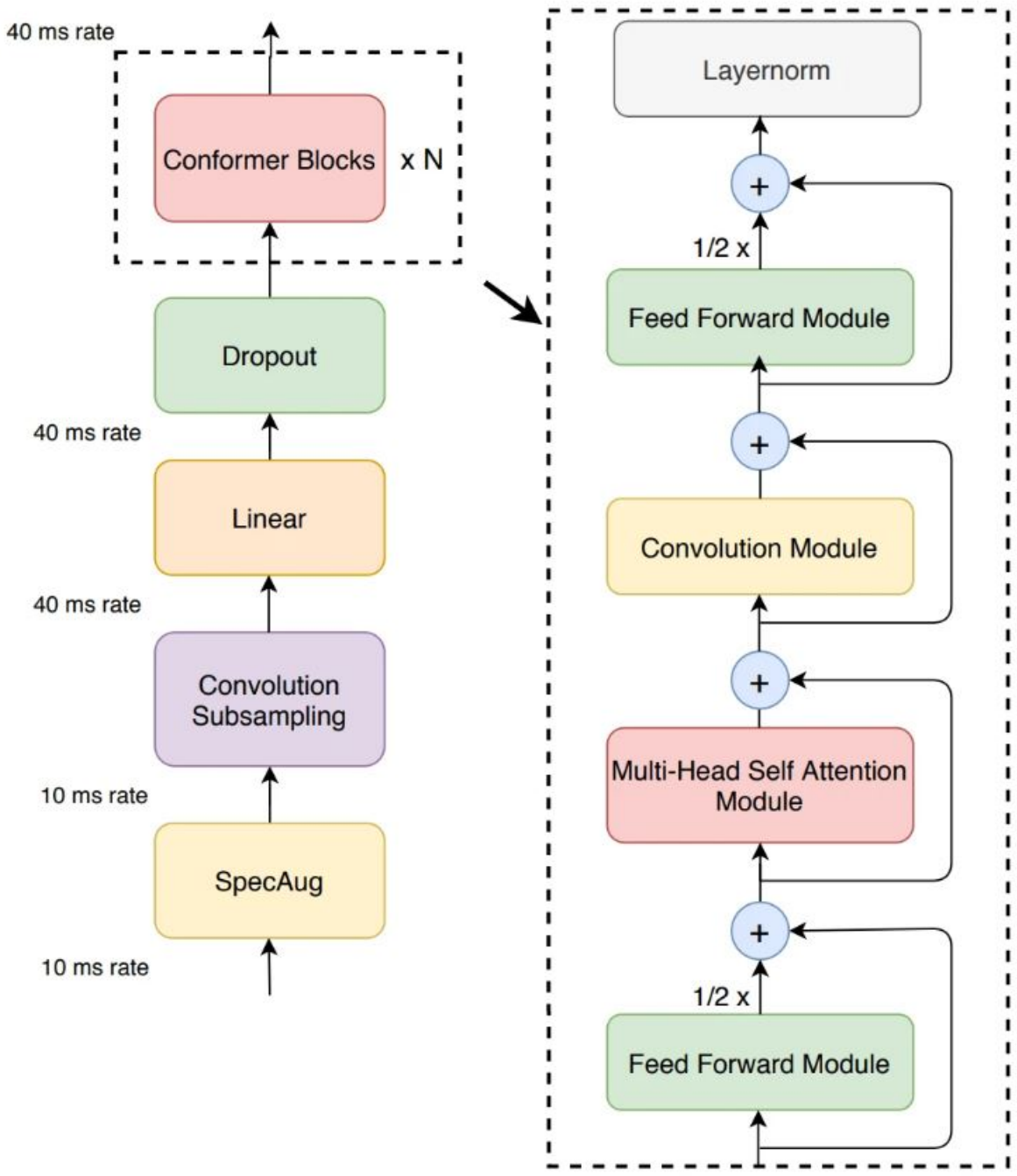

Fig. 4: Conformer CTC model architecture

## V. Aligning the Audio Books

***Manual Inspection*** left us with 8 audio books, each containing multiple <text, audio> pairs, instead of <single transcript, multiple audios> pairs. However, the problem of having long audios still persists. To divide 30minute long audios into many 7-8 second long audio chunk with corresponding texts, we used our baseline (obtained in IV) to get the alignments. We propose two strategies depending on the available resources and the duration of audios.

### A. *Neural Approach*

The **NeMo Forced Aligner (NFA)** matches text with spoken audio by leveraging token-level, word-level, and segment-level alignments. The common pipeline (Fig 5) uses a pretrained CTC model (**Conformer** in our case) to generate probabilistic outputs (logits) to perform soft alignment on the CTC side.

[4] No LLMs - replacing all the occurrences of "եւ" by "և"

[5] HF repo for our model - stt_arm_conformer_ctc_large

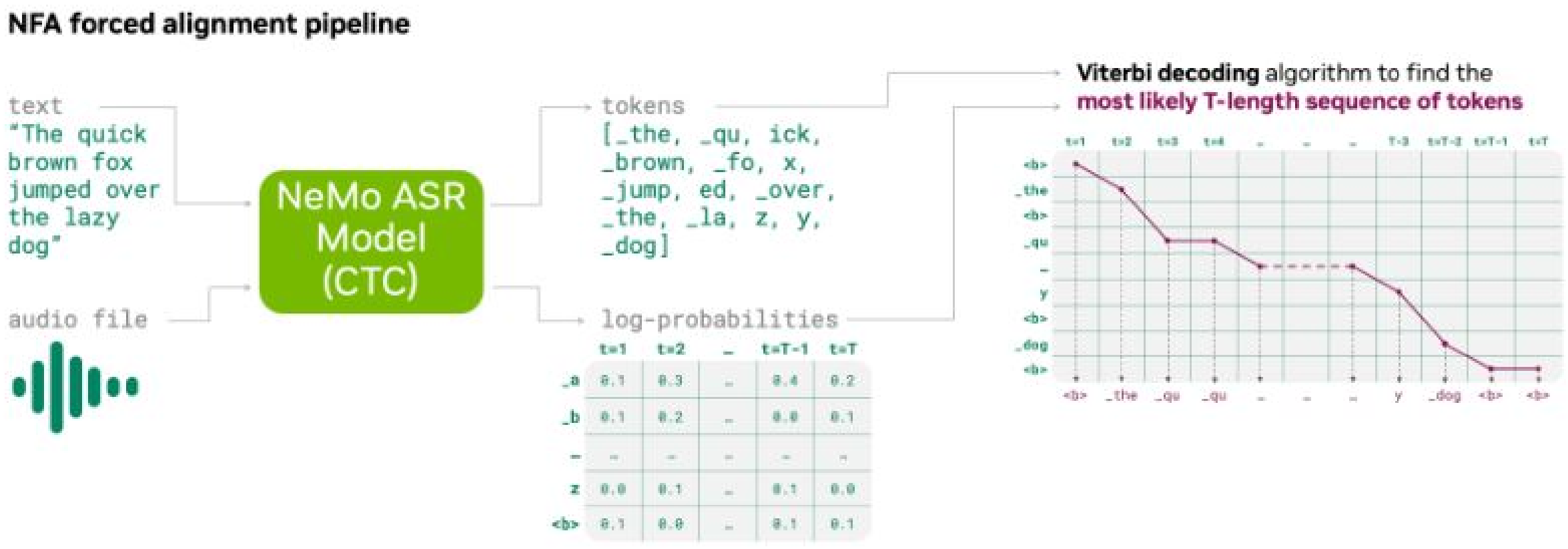

Fig. 5: NeMo-Forced-Aligner using a baseline model

As this paper aims to advance ASR for low-resource languages through dataset creation, we acknowledge the existence of rare cases where neither an available ASR model nor an MCV dataset is present, thus we suggest using tools like **MFA**[6] can be used.

After running NFA over the audio books corpus, we get segment, token and word level alignments - **.ctm** file (see example in Table III).

| Utterance | Start (s) | Duration(s) | Text |
|---|---|---|---|
| 001 | 0.04 | 2.12 | Ալեքսանդր |
| 001 | 2.20 | 0.76 | Ծիրվանզադե |
| 001 | 3.64 | 0.44 | Պատվի |
| 001 | 4.08 | 0.36 | Համար |
| 001 | 5.08 | 0.52 | դրամա |
| 001 | 5.76 | 0.36 | չորս |
| 001 | 6.20 | 0.80 | գործողությամբ |

TABLE III: Example of Word Level CTM Output

The word-level alignments, while accurate, are too brief — we aim for audio segments ranging from 3 to 15 seconds, targetting 8 seconds. To achieve this, we propose a grammar and duration-based segmentation algorithm to create longer chunks from these word-level alignments. We refer to this method as our **Segmentation Pipeline (SP)**:

1) **Sentence Boundary Detection:** We merge short word segments until they reach sentence-ending punctuation marks (e.g., '։' or ':' in Armenian). This step is crucial for identifying complete utterance units, which are typically comprised of a single, full sentence.

[6]Montreal Forced Aligner uses pre-trained acoustic models and phonetic dictionaries to align phonetic and word-level transcripts with audio recordings

2) **Error Mitigation in Segment Merging:** During the merging process, we address ASR errors that often lead to inaccurate end-time predictions of audio segments during the **NFA**. We use a simple **Voice Activity Detection (VAD)**, with an empirical threshold of **-20dB** tailored to our dataset, to identify abrupt endings or segments that terminate mid-word. Upon detection, the segment's end frame is adjusted to the closest point of silence or inactivity before proceeding with the merge.
3) **Splitting Long Segments:** Due to complex punctuation in audio-books, our segments sometimes extend up to 30 seconds. Therefore, we split them at **Auxiliary Punctuation (AP)**, such as commas. However, when faced with sentences that contain multiple APs, we greedily determine the most effective splits, by prioritizing the duration of silences around the APs while also minimizing the number of splits needed to obtain less than **15second** chunks. The objective is twofold: to ensure getting **complete voice utterance** and to reduce the frequency of cuts, thereby **maintaining the natural flow** of speech.
4) **Segment Consolidation:** Short segments that are contextually complete sentences, such as **'Բարև Ձեզ'** ('Hello.'), are merged with subsequent segments provided this does not breach the maximum segment duration constraint of 15seconds.
5) **Silence Trimming:** Excessive silences resulting from merging small chunks are trimmed to maintain the natural flow of speech.
6) **Concat Smoothing**: To ensure seamless tran-

sitions between segments and mitigate perceptible volume shifts, we apply modified Hanning Window at the junctions of merged segments.

7) **ASR Filtering:** We run our baseline model over the chunks, filtering out predictions with a WER above 0.75 to avoid potential error propagation and artifacts from NFA.

Finally, we obtain a new dataset with over 20hours of data (Fig 6). Despite having longer (close to 15seconds) audios in our new dataset compared to that of the MCV, data is still very valuable for ASR.

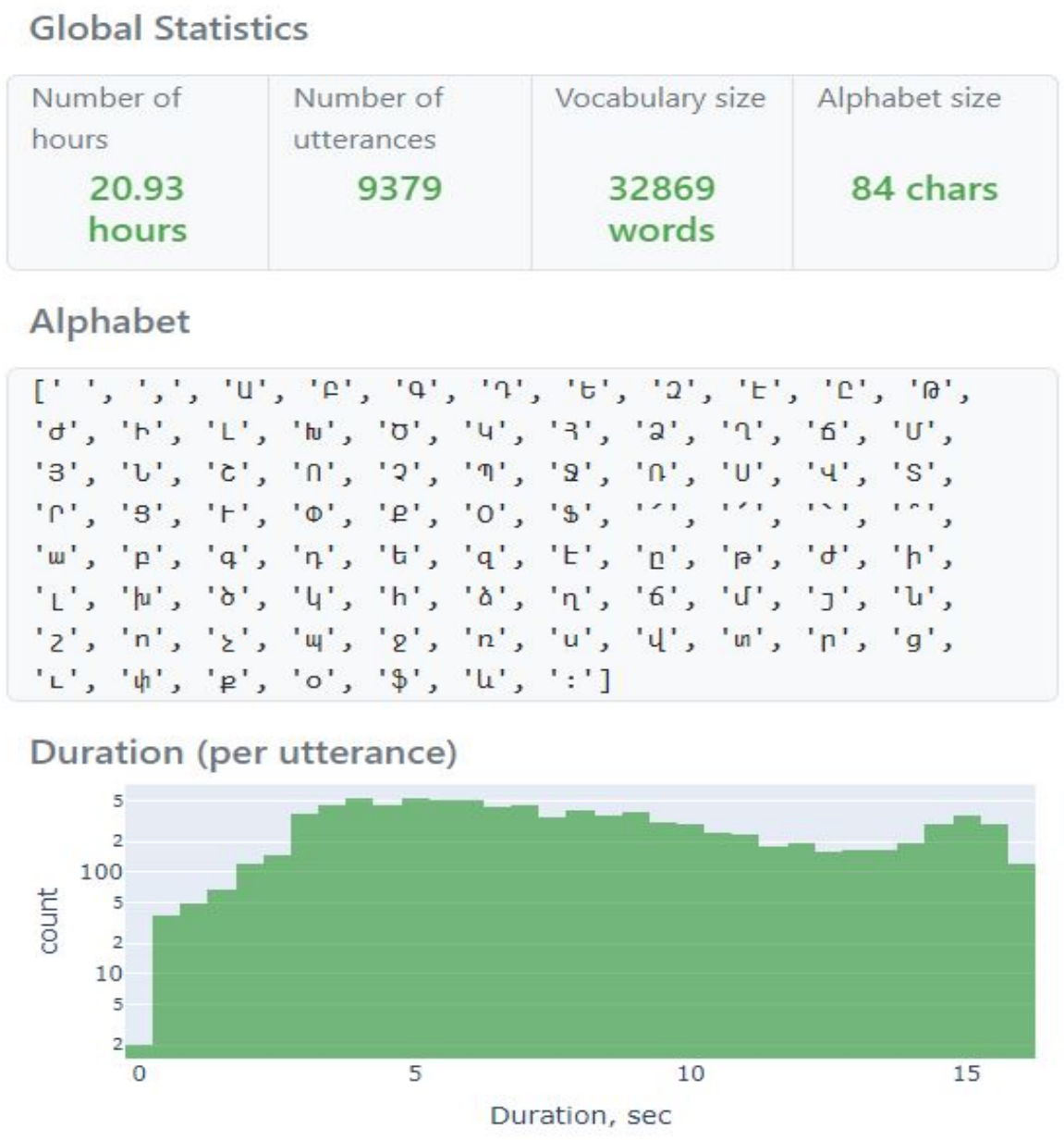

Fig. 6: Post NFA Audio Book (chunked) **v0**

The multi-step **SP** can be implemented in a more straightforward way - placing a special **splitter** token **"|"** after each End of Sentence punctuation in the raw transcripts. This will direct the **NFA** on its segment-level alignment (see IV). This is analogues to our SP's step 1) but with possibility of obtaining too long segments, which our SP cuts down at various levels.

| Start(s) | Duration (s) | Text Segment |
|---|---|---|
| 177.08 | 1.24 | Երիտասարդ: |
| 178.40 | 1.76 | Խոսում<space>է <space>Ծամախու <space>բարբառով: |
| 181.12 | 2.96 | Երիտասարդ <space>աղախին: |

TABLE IV: Example of Segment Level CTM Output

## B. ***Hybrid Approach - VAD-ASR-CER (VAC)***

Section V-A identifies **NFA** as the best practice for the Alignment task. However, its effectiveness comes with certain limitations. Firstly, the success of NFA significantly depends on the quality of the ASR baseline, which requires a substantial amount of data to achieve a close to 0.2 WER. Secondly, using NFA on audio longer than 30 minutes, even with high-capacity V100 GPUs (24Gb), greatly increases the likelihood of encountering memory allocation errors (see the **GitHub issue**). Given these challenges, we propose a novel method for aligning audio of **arbitrary** length, thus handling extensive multi-hour audios. Therefore, even Google Colab's free T4 GPUs are enough to run it. The pipeline[7] consists of:

1) **Voice Activity Detection** We split original audios based on the silences in the audios. This can results in having hundreds of up to 2second long chunks per 20-minute audio.
2) **ASR inference** Run baseline model over this chunks to obtain the corresponding texts.
3) **Character Error Rate (CER) matching**: Our complex algorithm iterates over the original transcript with a dynamic window, matching the source text with the predicted chunks' texts, while minimizing the CER.

**VAD-ASR-CER*** approach is independent of the audio length and even maintains its effectiveness with bad ASR models. Table V illustrates our capability of aligning poorly predicted texts. For a comprehensive understanding of the operational nuances of VAC, we highly recommend reviewing our **tutorials** as well as the detailed breakdown of VAC's 3rd and the **main** part (**Matching**) in Section X

| Source Text | Chunk Text | CER | Start | End |
|---|---|---|---|---|
| Գոլլում... Իսկ | Որ ո ի ում սկ: | 0.66 | 22918 | 22931 |
| միստր Բեգինս: | Մի ս Է բԲեկինձ: | 0.41 | 23072 | 23086 |
| Իսկ ո՞ւր է Տորինը | Ի՛սկ ՞ր է տօրինը: | 0.43 | 23272 | 23290 |

TABLE V: The **VAD-ASR-CER*** found matches (source text) for "The hobbit, or there and back again"

The second column displays ASR-predicted text, obtained after step 1) - **VAD**. The algorithm successfully identifies the **best match**, displayed in the "Source Text" column, corresponding to segments from the original transcript, in this case, from the book "The Hobbit and Back Again." The "Start" and "End" columns indicate the indices of these matched text segments in the originalraw transcript.

[7] Published PyPi Package: vac_aligner

These results of VAD-ASR-CER* can again be passed through our **Segmentation Pipeline** to obtain a final-grained dataset. Comparing to Section V-A VAC presents several advantages, particularly in the context of resource limitations. Primarily, it circumvents the need for extensive GPU resources required by NFAs during the alignment phase of lengthy audio files. Additionally, it alleviates the requirement for high-quality ASR models, which are often unfeasible in low-resource language scenarios due to data scarcity.

Despite these benefits, VAC relies heavily on the accuracy of the label transcript. Our algorithm operates by iteratively scanning the original, long-form transcript in sync with the post-VAD audio chunks. We dynamically adjusts the size and position of the text window to find the best match for each chunk's ASR-predicted labels, and continue this process throughout the transcript. However, this sequential and iterative method faces significant challenges if the transcript contains missing or extraneous text that could span a couple of sentences. For instance, introductory remarks at the beginning (sometimes at the end) of recordings, which often include date, chapter information, or other preliminary details not present in the transcript, can disrupt the matching process which could carry on throughout the whole transcript, resulting in wrong alignments. Consequently, the algorithm's performance deteriorates when encountering incomplete <audio, text> pairs. On the other side this is not a trouble maker as it is an expected outcome for having sub-optimal data.

To conclude the **VAC**, we had to perform some benchmarks on the algorithm. Initially, manual checks were essential, as predictions were occasionally nonsensical. Reviewing the matched text required examining a more extended context from the source text and sometimes listening to the corresponding audio. On average, the accuracy of the matches was 90.3% across 20 runs for each book. It is important to note that a 'false match' typically involves minor discrepancies, such as a missed or an additional word compared to the original transcript, which is a less critical error for the ASR than, for example, mislabeling an image of a dog as a table for the image classification task.

While these semi-quantitative measures yield valuable insights, a fully quantitative approach is essential to comprehensively evaluate the performance of our VAC pipeline. To this end, we devised a test scenario that closely mimics the audiobook data format. Specifically, we consolidated all texts from the MCV v17 test set into a single, extensive transcript to replicate the structure of an audiobook. We then ran our baseline model across all test audios and applied our **CER-based matching algorithm** to align the ASR predictions with this unified transcript, effectively identifying the correct source texts. The results are very promising, demonstrated in Table VI:

| Method | %Exact Matches | Mean WER | Mean CER |
|---|---|---|---|
| ASR only | 71% | 19% | 3% |
| VAC | **97%** | 0.5% | 0.34% |

TABLE VI: The quantitative benchmark for **VAC** on Merged MCV v17 Test Corpus

## VI. Retraining (Mixed dataset)

So far we did the following two important things:

1) **Baseline Model**: We conducted numerous training experiments on the fully-processed Mozilla Common Voice (MCV) v16 dataset to fine-tune hyper-parameters. We then transferred the refined model parameters for training on MCV v17 extended dataset. The outcome was a robust baseline model with a promising WER of 0.19, setting a strong foundation for subsequent enhancements.
2) **Processing and Enhancing Audio Books Data**: Initially unsuitable for training purposes, raw Audio Books data underwent a meticulous processing phase where we eventually aligned and then re-segmented the audio segments using our proposed algorithms. This effort resulted in a training set of slightly over 20 and a test set of over 1 hour, which can be easily extended having all the tool sets developed.

Building on these achievements, we are now poised to integrate the two distinct sources of data—totaling nearly 60 hours (after our Data Processing Pipeline - **54hours**)—to retrain our model. This phase presents unique challenges, as the datasets differ significantly in structure and quality. To ensure a successful merge and to avoid potential pitfalls during training, we have implemented several critical processing techniques, described below:

1) **Distributional Differences:**
   a) **Problem:** The Zero-Crossing Ratio and Root Mean Square Energy levels in the audio books are significantly lower than those observed in the Mozilla Common Voice (MCV) dataset. This discrepancy arises primarily due to extended silences within the audiobook recordings, occurring because of shifts between direct and indirect speech.

b) **Solution:** Reusing our VAD algorithm to cut these silences.

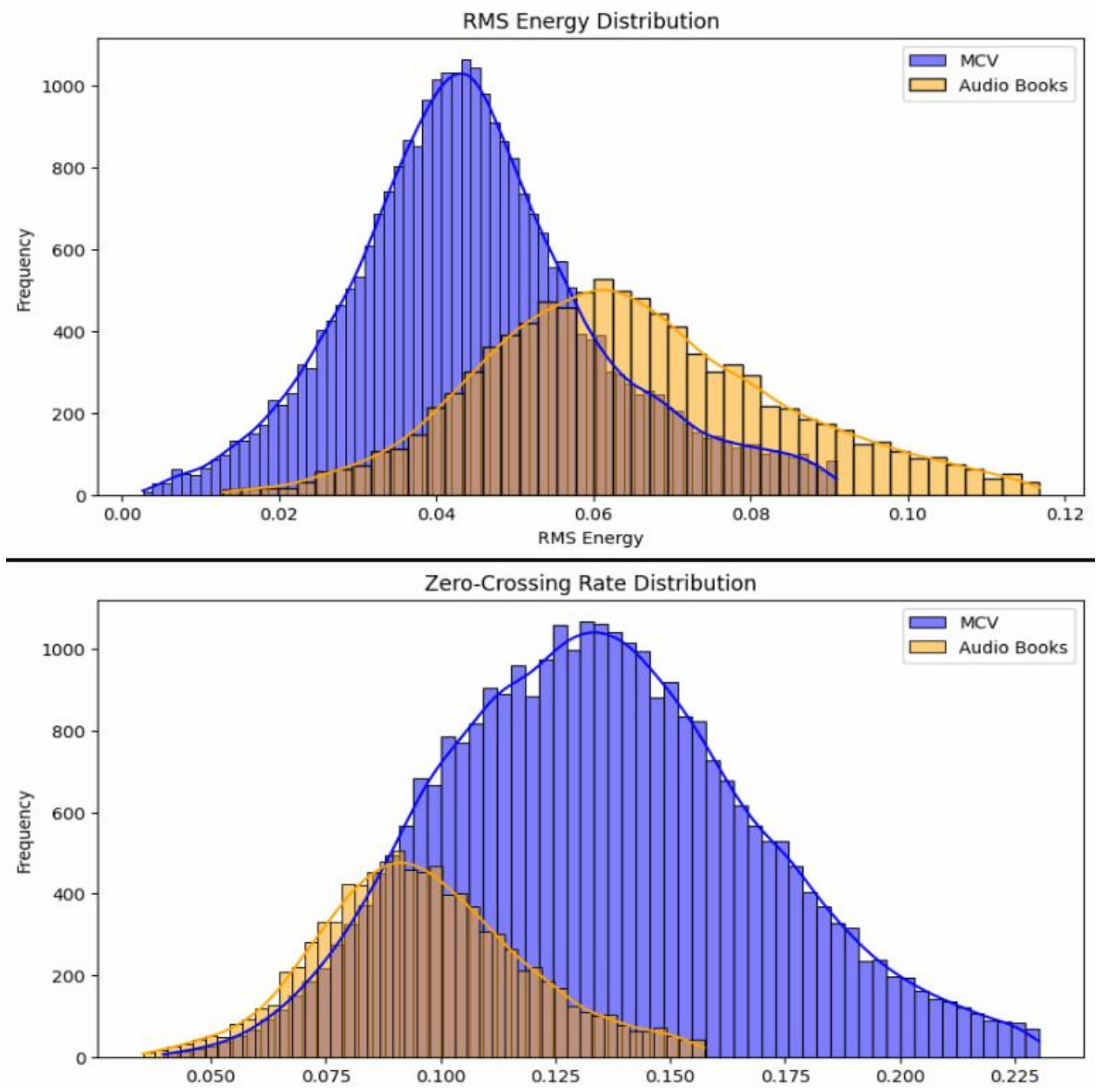

Fig. 7: Effect of Silences

2) **Speaking Style Differences:**
   a) **Problem:** MCV is a multi-speaker dataset, where each audio contains a single speaker, whereas audiobooks typically feature a single speaker per book. However, notable shifts in prosody and intonation occur within single audio chunks as the narrator attempts to mimic the voices of various characters during dialogues.
   b) **Solution:** Filter out audio segments that contain excessive use of Armenian punctuation marks "՜" or "՞", which are clear indicators of overly enthusiastic and expressive speech. This helps maintain a level of neutrality within the data.
3) **Volume Discrepancies Between Datasets:**
   a) **Problem:** A significant challenge we faced was the difference in volume levels between the MCV and the audio books. None of our initial training experiments with the mixed datasets yielded any notable results because of this huge gap.
   b) **Solution:** We applied a dynamic volume normalization across all audio files. This crucial step was imperative to ensure consistency in audio input levels, which helped to advance the training, underscoring the essential role of this process in the success of our model development.

## VII. Final Evaluation

The evaluation of our baseline and mixed training models offers intriguing insights into the effectiveness of different training data configurations for ASR systems. Initially, the baseline model, which was trained solely on 38 hours of the MCV v17 dataset, achieved a Word Error Rate (WER) of 0.19. However, when the model was trained on a mixed dataset comprising MCV v17 and an additional 20 hours from audiobooks, and tested on the same MCV test set (6.5 hours), the WER slightly increased to 0.21. This initial result suggested that the mixed training might be less effective on MCV-exclusive test data.

To further investigate, we evaluated both models on a test set composed solely of audiobook audios. The results were significantly better for the mixed model, which yielded a WER of 0.16, compared to the baseline's 0.39 WER. This substantial improvement highlights the mixed model's ability to generalize and capture underlying patterns in the Armenian language more effectively than the MCV-trained model. The nature of the MCV data, which consists of individually recorded sentences, differs markedly from the continuous, natural speech flow found in audiobooks. This continuous flow likely presents a more realistic scenario that an ASR system would encounter in production environments, suggesting why the mixed training model performed better on audiobook data.

| Training Data | Test Set | WER |
|---|---|---|
| MCV v17 only (38 hours) | MCV (6.5 hours) | 0.19 |
| MCV + Audiobooks (54 hours) | MCV (6.5 hours) | 0.21 |
| MCV v17 only (38 hours) | Audiobooks (1 hour) | 0.39 |
| MCV + Audiobooks (54 hours) | Audiobooks (1 hour) | 0.16 |

TABLE VII: Comparison of WER across different training and test configurations (*preserving both punctuation and capitalization*)

These results underscore the potential benefits of diversifying training data and adapting models to handle more natural speech patterns. Additionally, given the lengthy duration of each experiment due to the size of the extended training data, future efforts might focus on utilizing more extensive computational resources. Increasing the batch size and adjusting the learning rate warm-up steps, as discussed in our ablation study, could potentially enhance performance on the MCV test set as well.

## VIII. Future Work & Conclusions

Throughout this study, we have developed a novel pipeline to enhance ASR capabilities and demonstrated it on Armenian, a representative of low-resource languages. By leveraging audiobooks, we created rich, contextually varied ASR datasets, addressing significant gaps such as the scarcity of annotated speech data. Our approach introduces innovative algorithms like the Segmentation Pipeline for reassembling chunked audios and the **VAC (Voice Activity Detection-ASR-Character Error Rate)** matching system for aligning long audio files. These methodologies not only mitigate the data scarcity and resource constraints but are also scalable, making them applicable to other under-represented languages with accessible audiobooks. In addition, the data processing intuitions—considering the style and tone of audiobooks, the structural specifics of speech—and techniques like volume normalization and silence removal which were crucial in refining the training processes, are also applicable to audio books in other languages, as the core intuitions are shared. Furthermore, our hyperparameter optimization strategies to overcome initial bottlenecks in transfer learning (see IX) underscore the adaptability of our methods. Therefore, each component of our pipeline, from data processing to alignment and segmentation, from VAC to training, is designed to be portable and flexible across different languages, demonstrating our commitment to advancing ASR technology globally.

Looking forward, the potential to expand this work to other low-resource languages is vast. Specifically, the VAC pipeline can be leveraged to automate the transformation of audiobooks into training-ready data by replacing the **Manual Inspection** process (desc. in II-C). By merging all long audios into a single track and then applying the VAC, we can fully automate the data preparation process, from raw audio book data to chunked and segmented outputs suitable for model training. Future enhancements could also focus on refining the algorithm to better handle diverse dialects and accents, which pose significant challenges in low-resource settings. By improving the robustness of the VAC, our pipeline could achieve more accurate alignments, effectively handling segments where introductions or other text parts might be missing from the transcript. This capability is crucial given the frequent poor annotations in audiobook datasets, where long passages may lack precise textual correspondence, as highlighted in our methodology for processing audiobooks. The next big step would be experimenting on mixed datasets with various configurations, trying to obtain close to 0.1 WER on MCV dataset, fully beating the baseline. Simultaneously, we aim to maintain and enhance model generalization capabilities by evaluating performance on new test cases derived from audiobooks using our proposed pipeline.

## IX. Ablation Study

### A. *Introduction*

This chapter details an ablation study aimed at optimizing the EncDecModelBPE for Armenian speech recognition. We investigated several key hyperparameters, including batch size, learning rate schedulers, warm-up steps, and weight decay, to understand their effects on model performance. Our study leverages the MCV v16 and extends to version 17.

**Glossary:**

1) **Learning Rate (LR) Scheduler**: Defines how learning rate changes throughout the training
2) **Warm-up Steps**: Number of initial steps the learning rate increases to a predefined value
3) **Weight Decay:** Regularization technique to prevent overfitting by penalizing large weights

### B. *Importance of those hyper-parameters*

The pretrained conformer checkpoint[8], initially trained on a vast English corpus, is equipped with its own vocabulary and linguistic characteristics tailored to English. When we attempt to fine-tune this model for another language, we encounter significant **bottleneck**. This mismatch becomes apparent within just a few training steps, where the predictions initially comprising chaotic characters from our new tokenizer swiftly degenerate into sequences of empty strings. These issues persist unless we undertake specific optimizations of the hyper-parameters.

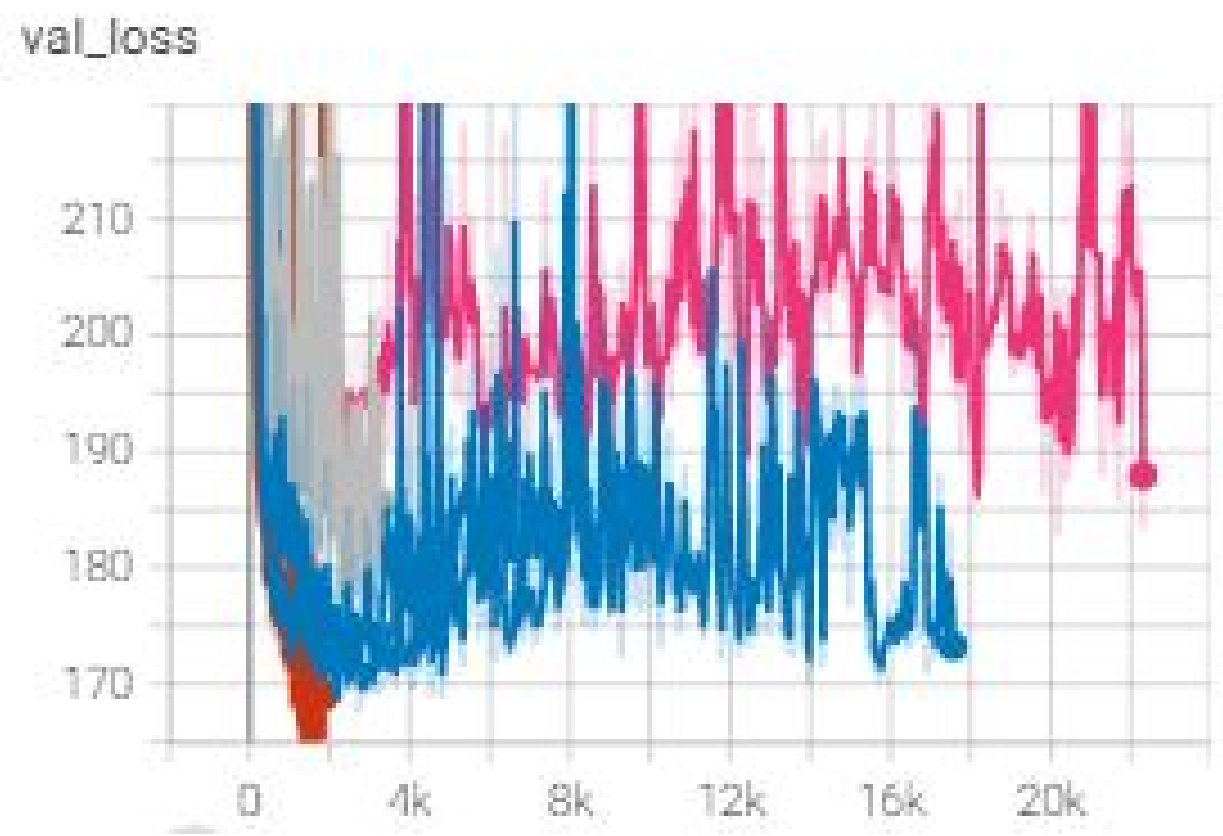

Fig. 8: The stucked training due to language transfer bottleneck. For reference, at least 20 validation loss is necessary to obtain a good model

[8]**stt_en_conformer_ctc_large** - https://catalog.ngc.nvidia.com/orgs/nvidia/teams/nemo/models/stt_en_conformer_ctc_large

A higher initial learning rate, facilitated by an appropriate scheduler (empirically - the best **NoamAnnealing**), provides the necessary momentum to overcome potential local minima early in training. This was visible both in the **stucked** validation plots (displayed in Fig 8), and validation logs, where we had predictions of whole utterances, consisting of only 1-2 characters, repeated over each other, when the scheduler strategy was for instance **Cosine Annealing**, which was lowering the learning rate too rapidly (Fig 11), preventing the model from effectively exploring the solution space.

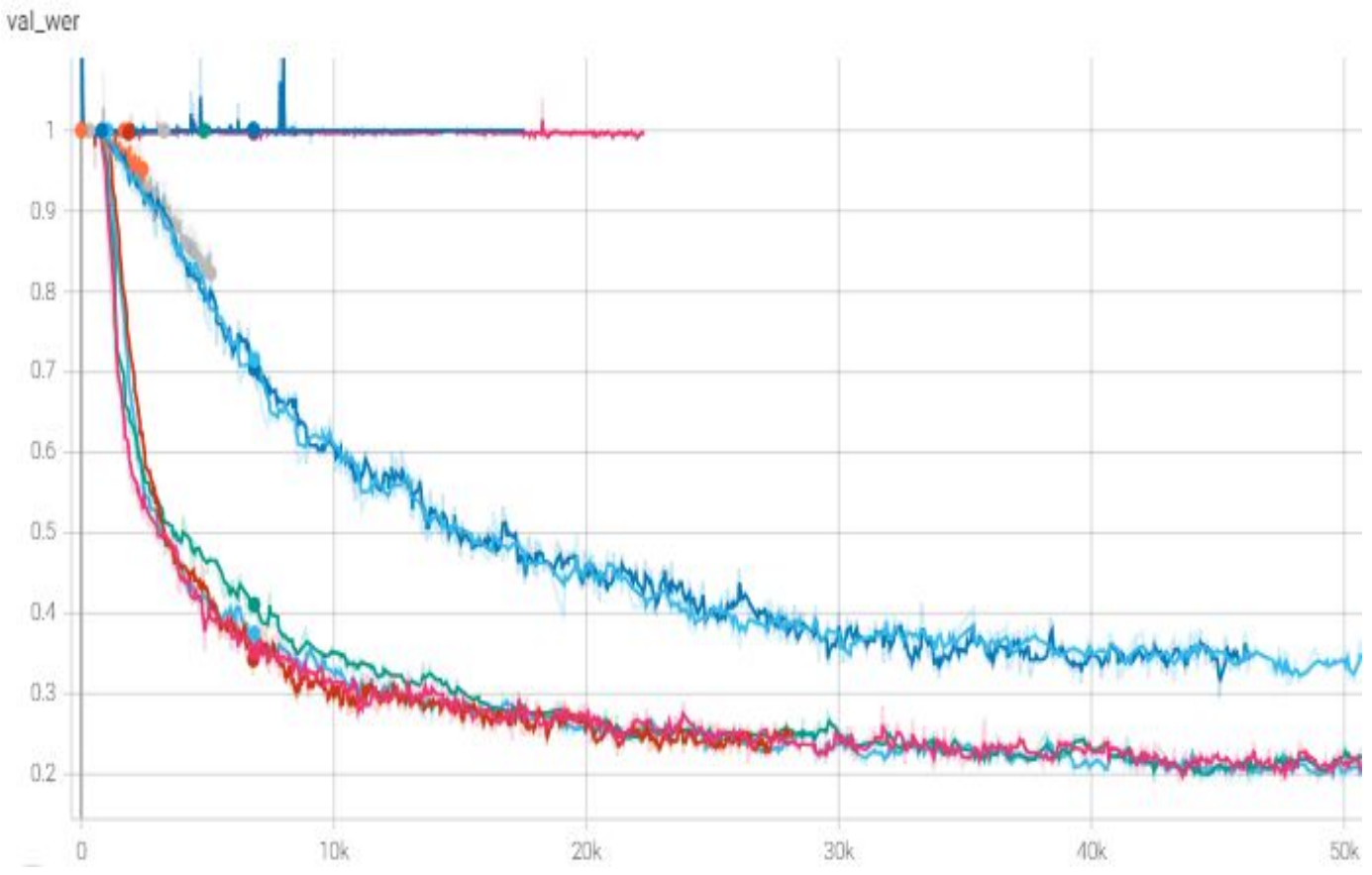

Fig. 9: Steeper Loss decrease when having higher learning rate and weight decay

Coming back to a higher initial learning rate, it proved beneficial for the model to quickly navigate the loss landscape, thereby avoiding potential shallow local minima that can derail the training process. This is supported by the observations in Fig 9, where models with higher learning rate have a significantly steeper decrease in loss, avoiding the prolonged horizontal asymptotes[9] that indicate stalled learning (around 1 WER lines in Fig 9). Though need to be cautious, as setting learning rate two high can cause both underfitting and overfitting (Fig 10).

Exploring the learning rate further, it is worh having a look at what actually this scheduling is visually.

[9]If looking at the training over 100k+ steps, can see "imaginable" horizontal asymtotes around which the **Validation WER** oscillates without any progress. This empiric observation helped to cut many non promising experiments

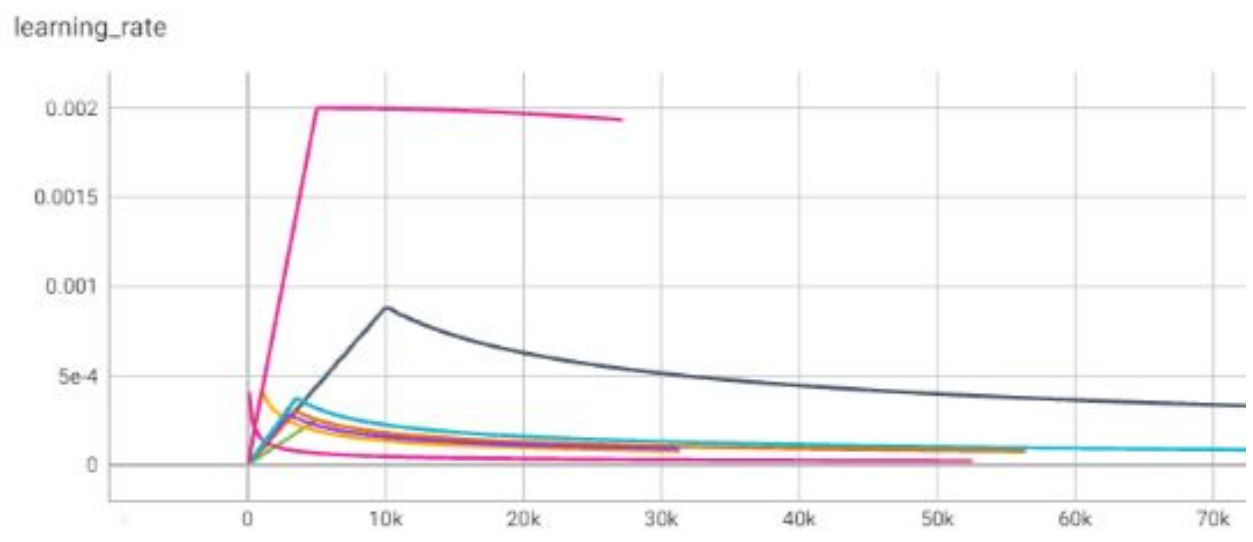

Fig. 11: The learning rate through training steps. The most aggressive (pink) is the same overfit training from Fig 10. The deviated in shape is the CosineAnnealing - quickly vanishing IX-B

### C. *Experimental Results*

Training configurations used over MCV v16 show us some patterns. The table below summarizes some configurations with their corresponding WERs, illustrating their influence on overall performance:

| **Warm-Up Steps** | **Learning Rate** | **Weight Decay** | **Test WER** |
|---|---|---|---|
| 10000 | 0.2 | 0.001 | 0.28 |
| 5000 | 0.4 | 0.4 | 0.25 |
| 3500 | 0.5 | 0.1 | **0.21** |
| 100 | 0.1 | - | 0.40 |

TABLE VIII: HParams on the MCV v16

These results underscore the effectiveness of a tailored approach to learning rate scheduling and hyperparameter tuning. Notably, the configuration with 3500 warm-up steps and a learning rate of 0.5 achieved the best WER of 0.21. Re-using them we managed to obtain 0.19 WER on the MCV v17 dataset. Moreover, using hard-code replacement we eventually obtained a 0.15 (0.12 w/o punctuation) WER which can be considered as open source Armenian SOTA model that we shared through **hugging face repo**.

During initial training trials, our model frequently settled into a counterproductive pattern, often predicting only outputting the colon (":"). This behavior was likely due to the colon's frequent presence at the end of sentences in the training data, leading the model to find a simplistic path to minimize loss. As a result, the model would rapidly converge to a loss asymptote and oscillate near this point, significantly hindering further learning and delaying meaningful results. To address this, we removing all the colons, from the texts. This adjustment discouraged the model from relying on punctuation prediction as a shortcut to reduce loss. While at inference time, to maintain the grammatical integrity of the generated text, we reintroduced colons where appropriate, especially if the sentence ended without any other punctuation such as a question mark. This post-processing step ensured that while the model was not biased towards predicting colons during training, the final output remained grammatically coherent.

Additionally, we implemented normalization techniques such as converting special Armenian names to lowercase (common pattern in audiobooks transcript, when during each persona's speehc author cites their names), which streamlined the training data and reduced the complexity that the model had to navigate. Such modifications were important as training on the mixed dataset inherently takes a couple of days.

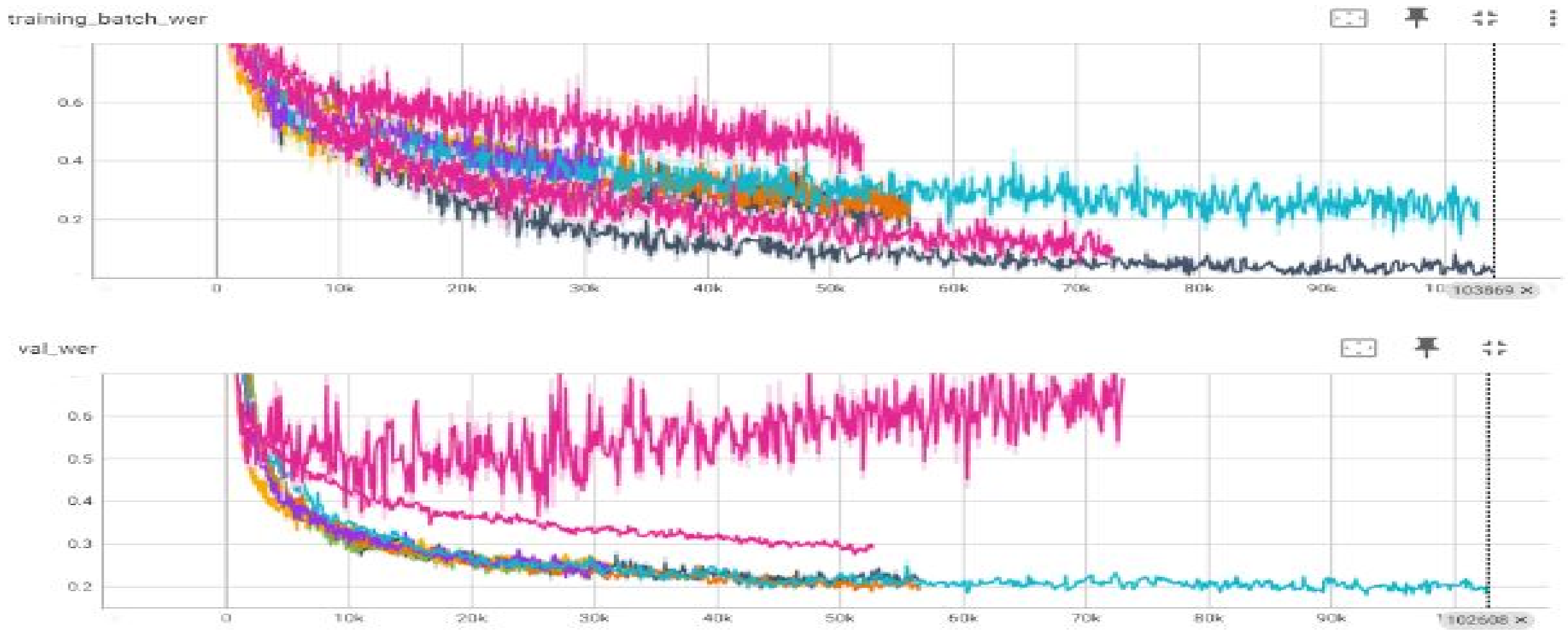

Fig. 10: The huge pink curve is an **overfit** (train decreases, val increases) with learning rate of 4, while middle one that is asymptotic towards WER 0.3 - is with learning rate of 2

## X. VAC Pipeline Breakdown

After the first 2 steps (relatively intuitive) of our pipeline - the VAD (chunking) and ASR (inference on those chunks), we have processed a long audio file and its corresponding transcript into multiple short audio chunks (each 3-15 seconds long) and their predicted texts from an Automatic Speech Recognition (ASR) model. These short audio segments are well-suited for ASR or Text-to-Speech (TTS) training. However, the predicted texts from the ASR model are not perfect and often contain errors such as repeated letters, incomplete words (e.g., "daw" instead of "draw"), or incorrect words (e.g., "dug" instead of "dog"). Therefore, it is necessary to align these predicted texts with the original transcript

### *A. Challenges of Matching Predicted Texts with the Original Transcript*

The ASR model's errors can lead to significant misalignments due to various factors such as repeated characters or words, incomplete words, and incorrect words that closely resemble the correct ones. These discrepancies often make a direct match between the original transcript and the predicted texts unfeasible, where even minor errors can lead to sequential mismatches and result in incomplete or redundant matches.

*Example Scenarios Illustrating Matching Challenges:*

1) The erroneous prediction results in incomplete extraction from the source text (missing or confusing some tokens spoken in the chunk audio):

   **Predicted Text ($\mathbf{i^{th}}$ chunk):** "Come here doggyyyy!!, dog, gggy, y?"

   **Predicted Text ($\mathbf{(i+1)^{th}}$ chunk):** "Are you serious?"

   **Original Transcript (context):** "... Come here doggy, doggy. Johnny, are you serious? Why did you hit the dog? ..."

   Here, if we naively take "*Come here doggy, doggy,*" from the source text as the best **CER-based** match (for the $i^{th}$ chunk), the next iteration will incorrectly include "*Johnny*" in the search, causing high CER and misalignment in future.

2) The erroneous prediction results in redundant extraction from the source text (mistakenly including some tokens not spoken in the $i^{th}$ chunk but in $(i+1)^{th}$):

   **Predicted Text ($\mathbf{i^{th}}$ chunk):** "Hey Madam, madam. Go mam"

   **Predicted Text ($\mathbf{(i+1)^{th}}$ chunk):** "Adam and Yeva are"

   **Original Transcript:** "... Hey Madam, madam. Adam and Yeva are ..."

   Here, if we naively take "*Hey Madam, madam. Adam,*" from the source text as the best **CER-based** match (for the $i^{th}$ chunk), the next iteration will miss "*Adam*" from the original source segment, causing further misalignments.

### *B. Proposed Solution: Dynamic (Greedy) Matching Algorithm*

To address the sequential nature and error-prone aspects of the matching process, we propose an advanced algorithm that attempts different combinations of source text segments, dynamically adjusts the search window, and greedily selects the best match at each step. Here is a high-level description of how it operates:

**Dynamic Window Adjustment:** The algorithm dynamically adjusts the size and position of the text window from the original transcript, referred to as the Source Segment (see examples from the previous Challenges Section). This adjustment helps align the predicted texts even when there are misalignments. For instance, correctly adjusting the start and end position of the Source Segment can help address errors from previous chunks (by extending the search window allows room to include missed tokens, while the CER threshold helps to remove redundant tokens).

**Greedy Selection:** At each step, the algorithm selects the best match based on similarity measures between the predicted text and the Source Segment. Similarity is defined using the Character Error Rate (CER), with a threshold of $\leq 0.3$. Although the algorithm is greedy in that it does not revisit matched texts once they are decided, it allows re-adjustment of the search window to cope with potential errors from previous matches, without breaking the sequential nature of the matches as the matches are not reconsidered.

**Sequential Error Mitigation:** The algorithm continuously mitigates sequential errors by refining its matches and avoiding the propagation of alignment errors. For example, if a match causes subsequent text segments to lag or overlap, the algorithm readjusts (extends its search window), allowing for correction in subsequent iterations.

*High-Level Example:* **Original Transcript:** “Once upon a time, in a faraway land, there lived a king.”

**Predicted Texts (chunks):**

1) “Once upon a tme”
2) “In a farway land”
3) “The’re livd a kng”

**Dynamic Matching Process:**

- **Initial Window:** The algorithm starts with a text window from the beginning of the original transcript. For the first chunk, it looks at a larger segment (e.g., “*Once upon a time, in a faraway land*”).
- **First Match:** The predicted text “*Once upon a tme*” is compared with the segment. The best match “*Once upon a time*” is found, and the algorithm moves to the next segment.
- **Adjust Window:** The window is adjusted to start from “a tme in a faraway land” - as there was a non-exact match because of the ASR error “tme” instead of “time”.
- **Subsequent Matches:** The process continues with the next predicted texts “*in a farway land*” and “the’re livd a kng”, adjusting the window each time based on the previous match.
- **Final Match:** The last segment “*the’re livd a kng*” is matched with “there lived a king”.

The algorithm ensures that each predicted text chunk aligns as closely as possible with the corresponding part of the original transcript, even if the matches are not exact. This dynamic and greedy approach helps in mitigating sequential errors and ensures a more accurate alignment overall.

| **Source Segment Match (post VAC)** | **ASR Prediction** | **CER** |
|---|---|---|
| Էն ի՞նչ է, ծառերի | Եմն ինչ է ն է: | 0.71 |
| էլի նորից: | Ե ՛ լ ին նորիցն: | 0.44 |
| դույզն իսկ չմտածելով՝ | Դուրսն ինձ չմտածելով.. | 0.36 |
| նույնը կրկնվեց, | Նա հին նակարգանվեց: | 0.71 |
| Դորի, Նորի, Օրի, Օյն, | Նուրորրի՛, նորի, օրի՞.. | 0.72 |
| դեռ «Բիլբոն» էլ ավելացնել | Կեռ բիլբոններր ավելացնելով: | 0.45 |
| տածեց Բիլբոն՝ | տածել Ֆիլմը: | 0.5 |
| — Գոլլում... Իսկ | Որ ո ի ում ն է: | 0.66 |
| միստր Բեգինս: | Մի ս է բԲեկինձ: | 0.41 |
| Իսկ դուք, ինչ է, | -ս դուք իհնչ է: | 0.38 |
| մի՛ ամաչիր... | Մի ամաչի՞ էր: | 0.5 |

TABLE IX: Demonstration of complex cases that VAC manages to match even with such poor ASR predictions